# Ecological Evaluation of Persuasive Messages Using Google AdWords

**Marco Guerini**
Trento-Rise
Via Sommarive 18, Povo
Trento — Italy
marco.guerini@trentorise.eu

**Carlo Strapparava**
FBK-Irst
Via Sommarive 18, Povo
Trento — Italy
strappa@fbk.eu

**Oliviero Stock**
FBK-Irst
Via Sommarive 18, Povo
Trento — Italy
stock@fbk.eu

## Abstract

In recent years there has been a growing interest in crowdsourcing methodologies to be used in experimental research for NLP tasks. In particular, evaluation of systems and theories about persuasion is difficult to accommodate within existing frameworks. In this paper we present a new cheap and fast methodology that allows fast experiment building and evaluation with fully-automated analysis at a low cost. The central idea is exploiting existing commercial tools for advertising on the web, such as Google AdWords, to measure message impact in an ecological setting. The paper includes a description of the approach, tips for how to use AdWords for scientific research, and results of pilot experiments on the impact of affective text variations which confirm the effectiveness of the approach.

*To appear at ACL 2012*

## 1 Introduction

In recent years there has been a growing interest in finding new cheap and fast methodologies to be used in experimental research, for, but not limited to, NLP tasks. In particular, approaches to NLP that rely on the use of web tools - for crowdsourcing long and tedious tasks - have emerged. Amazon Mechanical Turk, for example, has been used for collecting annotated data (Snow et al., 2008). However approaches *a la* Mechanical Turk might not be suitable for all tasks.

In this paper we focus on evaluating systems and theories about persuasion, see for example (Fogg, 2009) or the survey on persuasive NL generation studies in (Guerini et al., 2011a). Measuring the impact of a message is of paramount importance in this context, for example how affective text variations can alter the persuasive impact of a message.

The problem is that evaluation experiments represent a bottleneck: they are expensive and time consuming, and recruiting a high number of human participants is usually very difficult.

To overcome this bottleneck, we present a specific cheap and fast methodology to automatize large-scale evaluation campaigns. This methodology allows us to crowdsource experiments with thousands of subjects for a few euros in a few hours, by tweaking and using existing commercial tools for advertising on the web. In particular we make reference to the AdWords Campaign Experiment (ACE) tool provided within the Google AdWords suite. One important aspect of this tool is that it allows for real-time fully-automated data analysis to discover statistically significant phenomena. It is worth noting that this work originated in the need to evaluate the impact of short persuasive messages, so as to assess the effectiveness of different linguistic choices. Still, we believe that there is further potential for opening an interesting avenue for experimentally exploring other aspects of the wide field of pragmatics.

The paper is structured as follows: Section 2 discusses the main advantages of ecological approaches using Google ACE over traditional lab settings and state-of-the-art crowdsourcing methodologies. Section 3 presents the main AdWords features. Section 4 describes how AdWords features can be used for defining message persuasiveness metrics and what kind of stimulus characteristics can be evaluated. Finally Sections 5 and 6 describe how to build up an experimental scenario and some pilot studies to test the feasibility of our approach.

## 2 Advantages of Ecological Approaches

Evaluation of the effectiveness of persuasive systems is very expensive and time consuming, as the STOP experience showed (Reiter et al., 2003): designing the experiment, recruiting subjects, making them take part in the experiment, dispensing questionnaires, gathering and analyzing data.

Existing methodologies for evaluating persuasion are usually split in two main sets, depending on the setup and domain: (i) long-term, in the field evaluation of behavioral change (as the STOP example mentioned before), and (ii) lab settings for evaluating short-term effects, as in (Andrews et al., 2008). While in the first approach it is difficult to take into account the role of external events that can occur over long time spans, in the second there are still problems of recruiting subjects and of time consuming activities such as questionnaire gathering and processing.

In addition, sometimes carefully designed experiments can fail because: (i) effects are too subtle to be measured with a limited number of subjects or (ii) participants are not engaged enough by the task to provoke usable reactions, see for example what reported in (Van Der Sluis and Mellish, 2010). Especially the second point is awkward: in fact, subjects can actually be convinced by the message to which they are exposed, but if they feel they do not care, they may not "react" at all, which is the case in many artificial settings. To sum up, the main problems are:

1. Time consuming activities
2. Subject recruitment
3. Subject motivation
4. Subtle effects measurements

### 2.1 Partial Solution - Mechanical Turk

A recent trend for behavioral studies that is emerging is the use of Mechanical Turk (Mason and Suri, 2010) or similar tools to overcome part of these limitations - such as subject recruitment. Still we believe that this poses other problems in assessing behavioral changes, and, more generally, persuasion effects. In fact:

1. Studies must be as ecological as possible, i.e. conducted in real, even if controlled, scenarios.

2. Subjects should be neither aware of being observed, nor biased by external rewards.

In the case of Mechanical Turk for example, subjects are willingly undergoing a process of being tested on their skills (e.g. by performing annotation tasks). Cover stories can be used to soften this awareness effect, nonetheless the fact that subjects are being paid for performing the task renders the approach unfeasible for behavioral change studies. It is necessary that the only reason for behavior induction taking place during the experiment (filling a form, responding to a questionnaire, clicking on an item, etc.) is the exposition to the experimental stimuli, not the external reward. Moreover, Mechanical Turk is based on the notion of a "gold standard" to assess contributors reliability, but for studies concerned with persuasion it is almost impossible to define such a reference: there is no "right" action the contributor can perform, so there is no way to assess whether the subject is performing the action because induced to do so by the persuasive strategy, or just in order to receive money. On the aspect of how to handle subject reliability in coding tasks, see for example the method proposed in (Negri et al., 2010).

### 2.2 Proposed Solution - Targeted Ads on the Web

Ecological studies (e.g. using Google AdWords) offer a possible solution to the following problems:

1. Time consuming activities: apart from experimental design and setup, all the rest is automatically performed by the system. Experiments can yield results in a few hours as compared to several days/weeks.
2. Subject recruitment: the potential pool of subjects is the entire population of the web.
3. Subject motivation: ads can be targeted exactly to those persons that are, in that precise moment throughout the world, most interested in the topic of the experiment, and so potentially more prone to react.

4. Subject unaware, unbiased: subjects are totally unaware of being tested, testing is performed during their "natural" activity on the web.
5. Subtle effects measurements: if the are not enough subjects, just wait for more ads to be displayed, or focus on a subset of even more interested people.

Note that similar ecological approaches are beginning to be investigated: for example in (Aral and Walker, 2010) an approach to assessing the social effects of content features on an on-line community is presented. A previous approach that uses AdWords was presented in (Guerini et al., 2010), but it crowd-sourced only the running of the experiment, not data manipulation and analysis, and was not totally controlled for subject randomness.

## 3 AdWords Features

Google AdWords is Google's advertising program. The central idea is to let advertisers display their messages only to relevant audiences. This is done by means of keyword-based contextualization on the Google network, divided into:

- Search network: includes Google search pages, search sites and properties that display search results pages (SERPs), such as Froogle and Earthlink.
- Display network: includes news pages, topic-specific websites, blogs and other properties - such as Google Mail and The New York Times.

When a user enters a query like "cruise" in the Google search network, Google displays a variety of relevant pages, along with ads that link to cruise trip businesses. To be displayed, these ads must be associated with relevant keywords selected by the advertiser.

Every advertiser has an AdWords account that is structured like a pyramid: (i) account, (ii) campaign and (iii) ad group. In this paper we focus on ad groups. Each grouping gathers similar keywords together - for instance by a common theme - around an ad group. For each ad group, the advertiser sets a cost-per-click (CPC) bid. The CPC bid refers to the amount the advertiser is willing to pay for a click on his ad; the cost of the actual click instead is based on its quality score (a complex measure out of the scope of the present paper).

For every ad group there could be multiple ads to be served, and there are many AdWords measurements for identifying the performance of each single ad (its persuasiveness, from our point of view):

- CTR, Click Through Rate: measures the number of clicks divided by the number of impressions (i.e. the number of times an ad has been displayed in the Google Network).
- Conversion Rate: if someone clicks on an ad, and buys something on your site, that click is a conversion from a site visit to a sale. Conversion rate equals the number of conversions divided by the number of ad clicks.
- ROI: Other conversions can be page views or signups. By assigning a value to a conversion the resulting conversions represents a return on investment, or ROI.
- Google Analytics Tool: Google Analytics is a web analytics tool that gives insights into website traffic, like number of visited pages, time spent on the site, location of visitors, etc.

So far, we have been talking about text ads, - Google's most traditional and popular ad format - because they are the most useful for NLP analysis. In addition there is also the possibility of creating the following types of ads:

- Image (and animated) ads
- Video ads
- Local business ads
- Mobile ads

The above formats allow for a greater potential to investigate persuasive impact of messages (other than text-based) but their use is beyond the scope of the present paper[1].

## 4 The ACE Tool

AdWords can be used to design and develop various metrics for fast and *fully-automated* evaluation experiments, in particular using the ACE tool.

This tool - released in late 2010 - allows testing, from a marketing perspective, if any change made to a promotion campaign (e.g. a keyword bid) had a statistically measurable impact on the campaign itself. Our primary aim is slightly different: we are interested in testing how different messages impact (possibly different) audiences. Still the ACE tool goes exactly in the direction we aim at, since it incorporates *statistically significant testing* and allows avoiding many of the tweaking and tuning actions which were necessary before its release.

---
[1] For a thorough description of the AdWords tool see: https://support.google.com/adwords/

The ACE tool also introduces an option that was not possible before, that of *real-time testing* of statistical significance. This means that it is no longer necessary to define a-priori the sample size for the experiment: as soon as a meaningful statistically significant difference emerges, the experiment can be stopped.

Another advantage is that the statistical knowledge to evaluate the experiment is no longer necessary: the researcher can focus only on setting up proper experimental designs[2].

The limit of the ACE tool is that it only allows A/B testing (single split with one control and one experimental condition) so for experiments with more than two conditions or for particular experimental settings that do not fit with ACE testing boundaries (e.g. cross campaign comparisons) we suggest taking (Guerini et al., 2010) as a reference model, even if the experimental setting is less controlled (e.g. subject randomness is not equally guaranteed as with ACE).

Finally it should be noted that even if ACE allows only A/B testing, it permits the decomposition of almost any variable affecting a campaign experiment in its basic dimensions, and then to segment such dimensions according to control and experimental conditions. As an example of this powerful option, consider Tables 3 and 6 where control and experimental conditions are compared against every single keyword and every search network/ad position used for the experiments.

## 5 Evaluation and Targeting with ACE

Let us consider the design of an experiment with 2 conditions. First we create an ad Group with 2 competing messages (one message for each condition).

---

[2]Additional details about ACE features and statistics can be found at http://www.google.com/ads/innovations/ace.html

Then we choose the serving method (in our opinion the *rotate* option is better than *optimize*, since it guarantees subject randomness and is more transparent) and the context (language, network, etc.). Then we activate the ads and wait. As soon as data begins to be collected we can monitor the two conditions according to:

- Basic Metrics: the highest CTR measure indicates which message is best performing. It indicates which message has the highest initial impact.
- Google Analytics Metrics: measures how much the messages kept subjects on the site and how many pages have been viewed. Indicates interest/attitude generated in the subjects.
- Conversion Metrics: measures how much the messages converted subjects to the final goal. Indicates complete success of the persuasive message.
- ROI Metrics: by creating specific ROI values for every action the user performs on the landing page. The more relevant (from a persuasive point of view) the action the user performs, the higher the value we must assign to that action. In our view combined measurements are better: for example, there could be cases of messages with a lower CTR but a higher conversion rate.

Furthermore, AdWords allows very complex targeting options that can help in many different evaluation scenarios:

- Language (see how message impact can vary in different languages).
- Location (see how message impact can vary in different cultures sharing the same language).
- Keyword matching (see how message impact can vary with users having different interests).
- Placements (see how message impact can vary among people having different values - e.g. the same message displayed on Democrat or Republican web sites).
- Demographics (see how message impact can vary according to user gender and age).

### 5.1 Setting up an Experiment

To test the extent to which AdWords can be exploited, we focused on how to evaluate lexical variations of a message. In particular we were interested in gaining insights about a system for *affective* variations of existing commentaries on medieval frescoes for a mobile museum guide that attracts the attention of visitors towards specific paintings (Guerini et al., 2008; Guerini et al., 2011b). The various steps for setting up an experiment (or a series of experiments) are as follows:

**Choose a Partner**. If you have the opportunity to have a commercial partner that already has the infrastructure for experiments (website, products, etc.) many of the following steps can be skipped. We assume that this is not the case.

**Choose a scenario**. Since you may not be equipped with a VAT code (or with the commercial partner that furnishes the AdWords account and infrastructure), you may need to "invent something to promote" without any commercial aim. If a "social marketing" scenario is chosen you can select "personal" as a "tax status", that do not require a VAT code. In our case we selected cultural heritage promotion, in particular the frescoes of Torre Aquila ("Eagle Tower") in Trento. The tower contains a group of 11 frescoes named "Ciclo dei Mesi" (cycle of the months) that represent a unique example of non-religious medieval frescoes in Europe.

**Choose an appropriate keyword** on which to advertise, "medieval art" in our case. It is better to choose keywords with enough web traffic in order to speed up the experimental process. In our case the search volume for "medieval art" (in phrase match) was around 22.000 hits per month. Another suggestion is to restrict the matching modality on Keywords in order to have more control over the situations in which ads are displayed and to avoid possible extraneous effects (the order of control for matching modality is: *[exact match]*, *"phrase match"* and *broad match*).

Note that such a technical decision - which keyword to use - is better taken at an early stage of development because it affects the following steps.

**Write messages** optimized for that keyword (e.g. including it in the title or the body of the ad). Such optimization must be the same for control and experimental condition. The rest of the ad can be designed in such a way to meet control and experimental condition design (in our case a message with slightly affective terms and a similar message with more affectively loaded variations)

**Build an appropriate landing page**, according to the keyword and populate the website pages with relevant material. This is necessary to create a "credible environment" for users to interact with.

**Incorporate meaningful actions in the website**. Users can perform various actions on a site, and they can be monitored. The design should include actions that are meaningful indicators of persuasive effect/success of the message. In our case we decided to include some outbound links, representing:

- general interest: "Buonconsiglio Castle site"
- specific interest: "Eagle Tower history"
- activated action: "Timetable and venue"
- complete success: "Book a visit"

Furthermore, through new Google Analytics features, we set up a series of *time_spent_on_site* and *number_of_visited_pages* thresholds to be monitored in the ACE tool.

### 5.2 Tips for Planning an Experiment

There are variables, inherent in the Google AdWords mechanism, that from a research point of view we shall consider "extraneous". We now propose tips for controlling such extraneous variables.

**Add negative matching Keywords**: To add more control, if in doubt, put the words/expressions of the control and experimental conditions as negative keywords. This will prevent different highlighting between the two conditions that can bias the results. It is not strictly necessary since one can always control which queries triggered a click through the report menu. An example: if the only difference between control and experimental condition is the use of the adjectives "*gentle* knights" vs. "*valorous* knights", one can use two negative keyword matches: -*gentle* and -*valorous*. Obviously if you are using a keyword in exact matching to trigger your ads, such as *[knight]*, this is not necessary.

**Frequency capping for the display network**: if you are running ads on the display network, you can use the "frequency capping" option set to 1 to add more control to the experiment. In this way it is assured that ads are displayed only one time per user on the display network.

**Placement bids for the search network**: unfortunately this option is no longer available. Basically the option allowed to bid only for certain positions on the SERPs to avoid possible "extraneous variables effect" given by the position. This is best explained via an example: if, for whatever reason, one of the two ads gets repeatedly promoted to the premium position on the SERPs, then the CTR difference between ads would be strongly biased. From a research point of view "premium position" would then be an extraneous variable to be controlled (i.e. either both ads get an equal amount of premium position impressions, or both ads get no premium position at all). Otherwise the difference in CTR is determined by the "premium position" rather than by the independent variable under investigation (presence/absence of particular affective terms in the text ad). However even if it is not possible to rule out this "position effect" it is possible to monitor it by using the report (*Segment > Top vs. other + Experiment*) and checking how many times each ad appeared in a given position on the SERPs, and see if the ACE tool reports any statistical difference in the frequencies of ads positions.

**Extra experimental time**: While planning an experiment, you should also take into account the ads reviewing time that can take up to several days, in worst case scenarios. Note that when ads are in *eligible* status, they begin to show on the Google Network, but they are not approved yet. This means that the ads can only run on Google search pages and can only show for users who have turned off SafeSearch filtering, until they are approved. Eligible ads cannot run on the Display Network. This status will provide much less impressions than the final "approved" status.

**Avoid seasonal periods**: for the above reason, and to avoid extra costs due to high competition, avoid seasonal periods (e.g. Christmas time).

**Delivery method**: if you are planning to use the Accelerated Delivery method in order to get the results as quick as possible (in the case of "quick and dirty" experiments or "fast prototyping-evaluation cycles") you should consider monitoring your experiment more often (even several times per day) to avoid running out of budget during the day.

## 6 Experiments

We ran two pilot experiments to test how affective variations of existing texts alter their persuasive impact. In particular we were interested in gaining initial insights about an intelligent system for affective variations of existing commentaries on medieval frescoes.

We focused on adjective variations, using a slightly biased adjective for the control conditions and a strongly biased variation for the experimental condition. In these experiments we took it for granted that affective variations of a message work better than a neutral version (Van Der Sluis and Mellish, 2010), and we wanted to explore more finely grained tactics that involve the *grade of the variation* (i.e. a moderately positive variation vs. an extremely positive variation). Note that this is a more difficult task than the one proposed in (Van Der Sluis and Mellish, 2010), where they were testing long messages with lots of variations and with polarized conditions, neutral vs. biased. In addition we wanted to test how quickly experiments could be performed (two days versus the two week suggestion of Google).

Adjectives were chosen according to MAX bigram frequencies with the modified noun, using the Web 1T 5-gram corpus (Brants and Franz, 2006). Deciding whether this is the best metric for choosing adjectives to modify a noun or not (e.g. also pointwise mutual-information score can be used with a different rationale) is out of the scope of the present paper, but previous work has already used this approach (Whitehead and Cavedon, 2010). Top ranked adjectives were then manually ordered - according to affective weight - to choose the best one (we used a standard procedure using 3 annotators and a reconciliation phase for the final decision).

### 6.1 First Experiment

The first experiment lasted 48 hour with a total of 38 thousand subjects and a cost of 30 euros (see Table 1 for the complete description of the experimental setup). It was meant to test *broadly* how affective variations in the body of the ads performed. The two variations contained a fragment of a commentary of the museum guide; the control condition contained "*gentle* knight" and "*African* lion", while in the experimental condition the affective loaded variations were "*valorous* knight" and "*indomitable* lion" (see Figure 1, for the complete ads). As can be seen from Table 2, the experiment did not yield any significant result, if one looks at the overall analysis. But segmenting the results according to the keyword that triggered the ads (see Table 3) we discovered that on the "medieval art" keyword, the control condition performed better than the experimental one.

**Starting Date**: 1/2/2012  
**Ending Date**: 1/4/2012  
**Total Time**: 48 hours  
**Total Cost**: 30 euros  
**Subjects**: 38,082  
**Network**: Search and Display  
**Language**: English  
**Locations**: Australia; Canada; UK; US  
**KeyWords**: "medieval art", pictures middle ages  

Table 1: First Experiment Setup

| ACE split | Clicks | Impr. | CTR |
|---|---|---|---|
| Control | 31 | 18,463 | 0.17% |
| Experiment | 20 | 19,619 | 0.10% |
| Network | Clicks | Impr. | CTR |
| Search | 39 | 4,348 | 0.90% |
| Display | 12 | 34,027 | 0.04% |
| TOTAL | 51 | 38,082 | 0.13% |

Table 2: First Experiment Results

| Keyword | ACE split | Impr. | CTR |
|---|---|---|---|
| "medieval art" | Control | 657 | 0.76% |
| "medieval art" | Experiment | 701 | **0.14%**\* |
| medieval times history | Control | 239 | 1.67% |
| medieval times history | Experiment | 233 | 0.86% |
| pictures middle ages | Control | 1114 | 1.35% |
| pictures middle ages | Experiment | 1215 | 0.99% |

Table 3: First Experiment Results Detail. * indicates a statistically significant difference with $\alpha < 0.01$

**Discussion**. As already discussed, user motivation is a key element for success in such fine-grained experiments: while less focused keywords did not yield any statistically significant differences, the most specialized keyword "medieval art" was the one that yielded results (i.e. if we display messages like those in Figure 1, that are concerned with medieval art frescoes, only those users really interested in the topic show different reaction patterns to the affective variations, while those generically interested in medieval times behave similarly in the two conditions). In the following experiment we tried to see whether such variations have different effects when modifying a different element in the text.

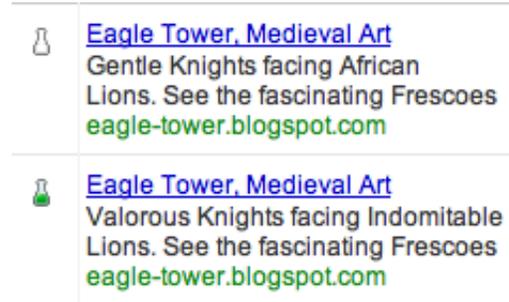

Figure 1: Ads used in the first experiment

## 6.2 Second Experiment

The second experiment lasted 48 hours with a total of one thousand subjects and a cost of 17 euros (see Table 4 for the description of the experimental setup). It was meant to test *broadly* how affective variations introduced in the title of the text Ads performed. The two variations were the same as in the first experiment for the control condition "*gentle* knight", and for the experimental condition "*valorous* knight" (see Figure 2 for the complete ads). As can be seen from Table 5, also in this case the experiment did not yield any significant result, if one looks at the overall analysis. But segmenting the results according to the search network that triggered the ads (see Table 6) we discovered that on the search partners at the "other" position, the control condition performed better than the experimental one. Unlike the first experiment, in this case we segmented according to the ad position and search network typology since we were running our experiment only on one keyword in exact match.

**Starting Date**: 1/7/2012  
**Ending Date**: 1/9/2012  
**Total Time**: 48 hours  
**Total Cost**: 17.5 euros  
**Subjects**: 986  
**Network**: Search  
**Language**: English  
**Locations**: Australia; Canada; UK; US  
**KeyWords**: [medieval knights]  

Table 4: Second Experiment Setup

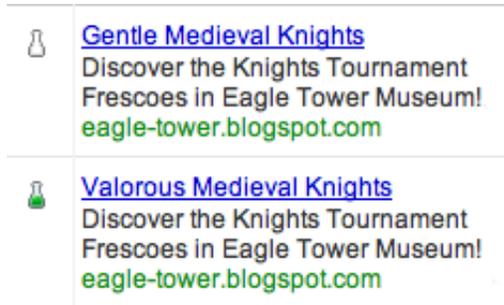

Figure 2: Ads used in the second experiment

| ACE split | Clicks | Impr. | CTR |
|---|---|---|---|
| Control | 10 | 462 | 2.16% |
| Experiment | 8 | **524**† | 1.52% |
| TOTAL | 18 | 986 | 1.82% |

Table 5: Second Experiment Results. † indicates a statistically significant difference with $\alpha < 0.05$

| Top vs. Other | ACE split | Impr. | CTR |
|---|---|---|---|
| Google search: Top | Control | 77 | 6.49% |
| Google search: Top | Experiment | 68 | 2.94% |
| Google search: Other | Control | 219 | 0.00% |
| Google search: Other | Experiment | **277*** | 0.36% |
| Search partners: Top | Control | 55 | 3.64% |
| Search partners: Top | Experiment | 65 | 6.15% |
| Search partners: Other | Control | 96 | 3.12% |
| Search partners: Other | Experiment | 105 | **0.95%**† |
| Total - Search | – | 986 | 1.82% |

Table 6: Second Experiment Results Detail. † indicates a statistical significance with $\alpha < 0.05$, * indicates a statistical significance with $\alpha < 0.01$

**Discussion**. From this experiment we can confirm that at least under some circumstances a mild affective variation performs better than a strong variation. This mild variations seems to work better when user attention is high (the difference emerged when ads are displayed in a non-prominent position). Furthermore it seems that modifying the title of the ad rather than the content yields better results: 0.9% vs. 1.83% CTR ($\chi^2$ = 6.24; 1 degree of freedom; $\alpha <$ 0,01) even if these results require further assessment with dedicated experiments.

As a side note, in this experiment we can see the problem of extraneous variables: according to AdWords' internal mechanisms, the experimental condition was displayed more often in the Google search Network on the "other" position (277 vs. 219 impressions - and overall 524 vs. 462), still from a research perspective this is not a interesting statistical difference, and ideally should not be present (i.e. ads should get an equal amount of impressions for each position).

## Conclusions and future work

AdWords gives us an appropriate context for evaluating persuasive messages. The advantages are fast experiment building and evaluation, fully-automated analysis, and low cost. By using keywords with a low CPC it is possible to run large-scale experiments for just a few euros. AdWords proved to be very accurate, flexible and fast, far beyond our expectations. We believe careful design of experiments will yield important results, which was unthinkable before this opportunity for studies on persuasion appeared.

The motivation for this work was exploration of the impact of short persuasive messages, so to assess the effectiveness of different linguistic choices. The experiments reported in this paper are illustrative examples of the method proposed and are concerned with the evaluation of the role of minimal affective variations of short expressions. But there is enormous further potential in the proposed approach to ecological crowdsourcing for NLP: for instance, different rhetorical techniques can be checked in practice with large audiences and fast feedback. The assessment of the effectiveness of a change in the title as opposed to the initial portion of the text body provides a useful indication: one can investigate if variations inside the *given* or the *new* part of an expression or in the *topic* vs. *comment* (Levinson, 1983) have different effects. We believe there is potential for a concrete extensive exploration of different linguistic theories in a way that was simply not realistic before.

## Acknowledgments

We would like to thank Enrique Alfonseca and Steve Barrett, from Google Labs, for valuable hints and discussion on AdWords features. The present work was partially supported by a Google Research Award.

## References

[Andrews et al.2008] P. Andrews, S. Manandhar, and M. De Boni. 2008. Argumentative human computer dialogue for automated persuasion. In *Proceedings of the 9th SIGdial Workshop on Discourse and Dialogue*, pages 138–147. Association for Computational Linguistics.

[Aral and Walker2010] S. Aral and D. Walker. 2010. Creating social contagion through viral product design: A randomized trial of peer influence in networks. In *Proceedings of the 31th Annual International Conference on Information Systems*.

[Brants and Franz2006] T. Brants and A. Franz. 2006. Web 1t 5-gram corpus version 1.1. *Linguistic Data Consortium*.

[Fogg2009] BJ Fogg. 2009. Creating persuasive technologies: An eight-step design process. *Proceedings of the 4th International Conference on Persuasive Technology*.

[Guerini et al.2008] M. Guerini, O. Stock, and C. Strapparava. 2008. Valentino: A tool for valence shifting of natural language texts. In *Proceedings of LREC 2008*, Marrakech, Morocco.

[Guerini et al.2010] M. Guerini, C. Strapparava, and O. Stock. 2010. Evaluation metrics for persuasive nlp with google adwords. In *Proceedings of LREC-2010*.

[Guerini et al.2011a] M. Guerini, O. Stock, M. Zancanaro, D.J. O'Keefe, I. Mazzotta, F. Rosis, I. Poggi, M.Y. Lim, and R. Aylett. 2011a. Approaches to verbal persuasion in intelligent user interfaces. *Emotion-Oriented Systems*, pages 559–584.

[Guerini et al.2011b] M. Guerini, C. Strapparava, and O. Stock. 2011b. Slanting existing text with Valentino. In *Proceedings of the 16th international conference on Intelligent user interfaces*, pages 439–440. ACM.

[Levinson1983] S.C. Levinson. 1983. *Pragmatics*. Cambridge University Press.

[Mason and Suri2010] W. Mason and S. Suri. 2010. Conducting behavioral research on amazon's mechanical turk. *Behavior Research Methods*, pages 1–23.

[Negri et al.2010] M. Negri, L. Bentivogli, Y. Mehdad, D. Giampiccolo, and A. Marchetti. 2010. Divide and conquer: Crowdsourcing the creation of cross-lingual textual entailment corpora. *Proc. of EMNLP 2011*.

[Reiter et al.2003] E. Reiter, R. Robertson, and L. Osman. 2003. Lesson from a failure: Generating tailored smoking cessation letters. *Artificial Intelligence*, 144:41–58.

[Snow et al.2008] R. Snow, B. O'Connor, D. Jurafsky, and A.Y. Ng. 2008. Cheap and fast—but is it good?: evaluating non-expert annotations for natural language tasks. In *Proceedings of the Conference on Empirical Methods in Natural Language Processing*, pages 254–263. Association for Computational Linguistics.

[Van Der Sluis and Mellish2010] I. Van Der Sluis and C. Mellish. 2010. Towards empirical evaluation of affective tactical nlg. In *Empirical methods in natural language generation*, pages 242–263. Springer-Verlag.

[Whitehead and Cavedon2010] S. Whitehead and L. Cavedon. 2010. Generating shifting sentiment for a conversational agent. In *Proceedings of the NAACL HLT 2010 Workshop on Computational Approaches to Analysis and Generation of Emotion in Text*, pages 89–97, Los Angeles, CA, June. Association for Computational Linguistics.